\documentclass[cameraready]{Interspeech}

\usepackage[utf8]{inputenc}
\usepackage[T1]{fontenc}
\usepackage{xcolor}
\usepackage[most]{tcolorbox}
\usepackage{enumitem} 
\usepackage{multirow}
\usepackage{amssymb}
\usepackage[table]{xcolor} 

\title{ROMPAR: Morphological Completion and Demographic Unlearning for Romanian-Accented Speech Recognition}

\author[affiliation={1}]{Andrei-Marius}{Avram}
\author[affiliation={1}, equalcontribution]{Aureliu-Valentin}{Antonie}
\author[affiliation={1}, equalcontribution]{Ştefan-Bogdan}{Badea}
\author[affiliation={1}, equalcontribution]{Andrei}{Florea}
\author[affiliation={1}, equalcontribution]{Robert-Nicolae}{Zaharoiu}
\author[affiliation={1}]{Dumitru-Clementin}{Cercel}

\address{
    $^1$ National University of Science and Technology POLITEHNICA Bucharest, Romania
}

\email{dumitru.cercel@upb.ro}


\usepackage{comment}

\begin{document}

\maketitle

\begin{abstract}

Automated transcription of parliamentary proceedings faces significant hurdles due to demographic bias, dialectal variation, and technical artifacts such as utterance truncation during segmentation. This paper introduces the ROManian PARliamentary Speech Corpus (ROMPAR) dataset, a 17.80-hour corpus of Romanian and Moldavian parliamentary speech, featuring double-annotated ground truth and explicit labels for reconstructed word fragments. To build a robust ASR system, we propose a multi-task adversarial training framework that enforces demographic invariance across age, gender, and dialect. We address the inherent instability of adversarial objectives in generative architectures by introducing an exponential decay mechanism for the adversarial coefficients. Furthermore, we implement an LLM-guided decoding strategy with position-dependent weighting to facilitate morphological completion of truncated terminal words. Our results demonstrate that the proposed framework significantly reduces WER and achieves an F1-score of 96.6\% in morphological reconstruction. 
\end{abstract}

\section{Introduction}

Automatic Speech Recognition (ASR) has witnessed a paradigm shift with the advent of large-scale generative models, achieving near-human performance in general-purpose domains. However, deploying these systems in specialized environments, such as legislative assemblies, presents a unique set of challenges. Parliamentary proceedings are characterized by a formal yet spontaneous speaking style, high-perplexity vocabulary, and often challenging acoustic conditions including reverberation and overlapping speech. For low-to-medium resource languages like Romanian, these difficulties are further compounded by significant dialectal variations—most notably between standard Romanian and the Moldavian dialect—and the demographic imbalances inherent to political institutions, which can bias models against underrepresented speaker groups.

A critical, yet often overlooked issue in the automated transcription of continuous legislative sessions is the problem of audio segmentation. Long-form parliamentary sessions are typically sliced into shorter utterances for processing; however, automated Voice Activity Detection (VAD) often truncates the final phonemes of a sentence due to hesitation or rapid turn-taking. This results in incomplete morphological structures where the acoustic evidence for the final suffix is absent. As illustrated in Figure \ref{fig:translation-comparison}, standard ASR models frequently fail to recover these endings (e.g., transcribing \textit{condi} instead of \textit{condi[țiile]}). Recovering this lost information—a task we term "morphological completion"—is essential for maintaining grammatical correctness and ensuring the semantic fidelity required for official public records.

\begin{figure}
    \centering
    \begin{tcolorbox}[
        enhanced,
        colback=white,
        colframe=gray!50!black, 
        boxrule=1pt,
        arc=4pt,
        left=4mm, right=4mm, top=4mm, bottom=4mm,
    ]
        {\color{blue!70!black}\sffamily\bfseries\large ROMANIAN/MOLDAVIAN}
        \vspace{1mm}
        \begin{itemize}[nosep, leftmargin=*]
            \item când va obține autorizația de demolare în condi{\color{red}[țiile]}...
            \item plaja de beneficiari de asigurări gratuite de sănătate și la elevi care învață în străinătate și la abso{\color{red}[lvenți]}
            \item să nu fie percepută ca și o intervenție punctuală, am avut această inter{\color{red}[venție]}
        \end{itemize}
    
        \vspace{1mm}
        \tcbline 
        \vspace{1mm}
    
        {\color{green!50!black}\sffamily\bfseries\large ENGLISH}
        \vspace{1mm}
        \begin{itemize}[nosep, leftmargin=*]
            \item when will the demolition permit be obtained under the condi{\color{red}[tions]}...
            \item the range of beneficiaries of free health insurance also includes students studying abroad and gradu{\color{red}[ates]}
            \item so that it is not perceived as a one-off intervention, we had this inter{\color{red}[vention]}
        \end{itemize}
    \end{tcolorbox}
\vspace{-2mm}
\caption{Comparative analysis of morphological completion in Romanian and English text samples.}
    \label{fig:translation-comparison}
\end{figure}

In this work, we address these multifaceted challenges by introducing a holistic framework for robust parliamentary ASR. We present the ROMPAR dataset, a rigorously curated corpus of Romanian and Moldavian legislative speech. To support the study of morphological completion, our dataset features a novel double-annotation protocol where truncated word endings are explicitly reconstructed and marked with bracketed notation, providing a ground truth for learning non-audible linguistic content.

To ensure our model remains robust across the diverse demographics of the parliament, we employ a multi-task adversarial training strategy. By reversing the gradient for auxiliary tasks—specifically age, gender, and dialect identification—we incentivize the encoder to discard speaker-specific acoustic signatures and focus solely on linguistic content. While adversarial learning is common in discriminative models, it is notoriously unstable in generative architectures. We propose a solution via an exponential decay mechanism for adversarial coefficients, which stabilizes the training dynamics and prevents decoder collapse. Finally, to address the truncation problem where acoustic cues are missing, we integrate an LLM-guided decoding strategy. By applying a position-dependent weight that boosts the language model's influence specifically on terminal tokens, we enable the system to "hallucinate" the correct morphological suffixes based on syntactic context.

Our primary contributions are summarized as follows:
\begin{itemize}
    \item \textbf{Dataset Release:} We introduce the first dataset for speech recognition across Romanian regional varieties from Romania and the Republic of Moldova. The ROMPAR dataset\footnote{Our corpus is freely available at: \url{https://huggingface.co/datasets/avramandrei/rompar}.} is a double-annotated open-sourced corpus, including metadata for dialect, age, and gender, alongside unique annotations for truncated word reconstruction.
    \item \textbf{Stabilized Adversarial Training:} We propose an exponential decay strategy for adversarial objectives, successfully enabling generative ASR models to learn dialect and demographic invariance without sacrificing transcription quality.
    \item \textbf{Morphological Completion:} We develop an LLM-guided decoding mechanism with terminal bracket weighting, which significantly improves the recovery of truncated words compared to standard decoding baselines.
\end{itemize}

\section{Related Work}

Automatic Speech Recognition (ASR) has undergone a significant paradigm shift, transitioning from traditional Hidden Markov Models (HMM) and manual feature extraction to deep learning-based end-to-end (E2E) architectures \cite{ahlawat2025automatic}. The field is currently dominated by Transformer \cite{vaswani2017attention} and Conformer \cite{gulati2020conformer} models, which utilize self-attention mechanisms to capture long-range dependencies and convolutional modules to extract local feature patterns. Recent advancements have been driven by foundational models such as Whisper \cite{peng2024owsm} and wav2vec 2.0 \cite{baevski2020wav2vec}, which leverage large-scale supervised or self-supervised pre-training to achieve robust performance across diverse and noisy environments.

Despite these global advancements, Romanian, especially the Moldavian dialect, remains a relatively under-resourced language, characterized by a scarcity of manually annotated data compared to high-resource languages. Recent efforts have expanded available Romanian resources through datasets like the Read Speech Corpus (RSC) \cite{georgescu2020rsc} and the Spontaneous Speech Corpus (SSC) \cite{georgescu2019progress}, with state-of-the-art performance now being achieved through the adaptation of efficient architectures like FastConformer \cite{pirlogeanu2025open}.

Adversarial methods have become increasingly vital for enhancing ASR robustness and fairness. Research has explored defenses against white-box adversarial attacks through joint adversarial fine-tuning with denoisers to protect models from imperceptible perturbations \cite{joshi2022defense}. Furthermore, Generative Adversarial Networks (GANs) are frequently employed for data augmentation, particularly to improve generalization in low-resource or disordered speech tasks \cite{wang2024enhancing}. Crucially, Domain Adversarial Training (DAT) is now being utilized to mitigate demographic biases by enforcing the learning of domain-invariant representations via gradient or loss reversal, thereby ensuring more equitable model performance across diverse speaker groups \cite{kim2025domain,avram2025rolargesum}.

\section{ROMPAR Dataset}
\label{sec:dataset}

The ROMPAR dataset consists of audio recordings collected from parliamentary proceedings in Romania and Moldova. This domain was chosen to ensure a rich vocabulary and a formal speaking style characteristic of legislative assemblies.

\subsection{Data Collection and Annotation}
The raw audio data was manually segmented and annotated by a team of 5 native speakers. To maximize the reliability of the ground truth, we employed a double-annotation strategy: each audio sample was processed by two independent annotators who provided orthographic transcriptions and metadata labels. The metadata includes the speaker's dialect, gender, and age group.

Due to the automated nature of the initial segmentation, a small portion of audio samples contained truncated words at the end of the recording. Annotators were instructed to reconstruct these missing endings based on context and enclose the inferred text in square brackets (e.g., Romanian: \textit{parlament[ar]}, translated to English as: \textit{parliament[ary]}). This protocol ensures grammatical consistency while explicitly marking non-audible phonetic content.

\subsection{Dataset Statistics}

The final corpus comprises a total of 17.80 hours of speech data across 14,891 samples. As detailed in Table \ref{tab:dataset_stats}, the data is partitioned into training (14.20 hours), validation (1.72 hours), and test (1.88 hours) sets. The transcriptions contain a total of 156,147 words, with an average sentence length of approximately 10.5 words per sample. To assess the acoustic environment, we calculated the Signal-to-Noise Ratio (SNR) and the Signal-to-Reverberation Ratio (SRR). The dataset exhibits a mean SNR of 21.05 dB and a mean SRR of 23.37 dB. These values reflect the high-fidelity recording equipment used in parliamentary chambers while accounting for the inherent background noise and acoustic reflections typical of large legislative halls.

\begin{table}
\centering
\caption{Dataset statistics showing the number of samples, total hours, average number of words, and total words in transcripts for each subset.}
\begin{tabular}{|l|cccc|}
\toprule
\multirow{2}{*}{\textbf{Subset}} & \multirow{2}{*}{\textbf{\#Samples}} & \multirow{2}{*}{\textbf{\#Hours}} & \textbf{Avg.} & \textbf{Total} \\
 & & & \textbf{Words} & \textbf{Words} \\
\midrule
Train & 11,912 & 14.20 & 10.4 & 124,410 \\
Valid & 1,489 & 1.72 & 10.2 & 15,144 \\
Test & 1,490 & 1.88 & 11.1 & 16,593 \\
\midrule
\textbf{Total} & \textbf{14,891} & \textbf{17.80} & \textbf{10.5} & \textbf{156,147} \\
\bottomrule
\end{tabular}
\label{tab:dataset_stats}
\end{table}

\begin{figure}
    \centering
    \includegraphics[width=1.0\linewidth]{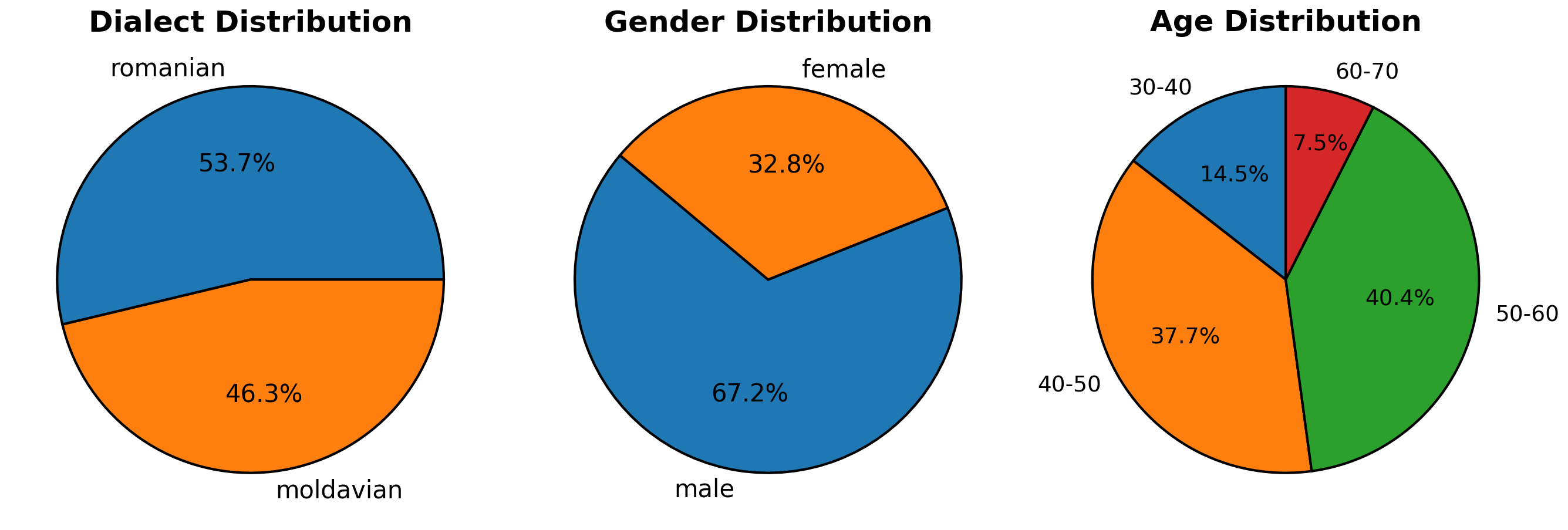}
    \caption{Demographic distribution of the ROMPAR dataset across three metadata categories: Dialect, Gender, and Age Group.}
    \label{fig:dataset_dist}
\end{figure}

Figure \ref{fig:dataset_dist} illustrates the demographic distribution of the corpus. The dataset exhibits a relatively balanced dialect representation, with 53.7\% of samples labeled as Romanian and 46.3\% as Moldavian. The gender distribution reflects the natural imbalance often found in parliamentary data, with 67.2\% male and 32.8\% female speakers. Regarding age, the majority of the speakers fall into the middle-aged categories: the 50-60 age group is the most represented (40.4\%), followed closely by the 40-50 group (37.7\%). Younger speakers (30-40) and seniors (60-70) account for 14.5\% and 7.5\% of the data, respectively.

\subsection{Inter-Annotator Agreement}

To assess the quality of the dataset, we evaluated the consistency between the two independent annotators. For the orthographic transcriptions, we computed the pairwise Word Error Rate (WER) between annotator outputs. The average pairwise WER was 3.4\%, indicating a high level of transcriptional accuracy. Disagreements were primarily related to punctuation and hesitation markers, which were resolved by a third senior linguist.

For the metadata labels, we measured agreement using Cohen's Kappa coefficient ($\kappa$) \cite{cohen1960coefficient}. The annotations showed almost perfect agreement for Gender ($\kappa = 0.98$) and Dialect ($\kappa = 0.96$). The agreement for Age groups was substantial ($\kappa = 0.87$), with minor confusion occurring only between adjacent age brackets (e.g., boundaries between 40-50 and 50-60).

\section{Methodology}
\label{sec:methodology}

Our approach aims to build a robust ASR system that is invariant to demographic variations—specifically age, dialect, and gender—while maintaining high transcription fidelity. We fine-tune various generative ASR models (i.e., Open Whisper \cite{peng2024owsm},
Granite Speech 3.3 \cite{saon2025granite}, Voxtral \cite{liu2025voxtral}, Granite Speech 3.3 \cite{yang2025qwen3},
and Parakeet TDT\cite{sekoyan2025canary}) using our multi-task adversarial framework. The models obtained in this work are marked with $^+$.

\subsection{Adversarial Training via Loss Reversal}

The primary objective is to minimize the ASR loss, $\mathcal{L}_{\text{asr}}$, while simultaneously forcing the model to learn features that are uninformative for demographic classification. Unlike standard multi-task learning where all losses are minimized, we employ an adversarial objective for the demographic classifiers $D \in \{\text{age, dialect, gender}\}$. 

Specifically, we reverse the sign of the adversarial components in our total loss function, as introduced in \cite{avram2024histnero}. The joint objective $\mathcal{L}_{\text{total}}$ is formulated as:

\begin{equation}
    \mathcal{L}_{\text{total}} = \mathcal{L}_{\text{ASR}} - \sum_{d \in D} \lambda_d(t) \cdot \mathcal{L}_{d}
\end{equation}

where $\mathcal{L}_{d}$ represents the cross-entropy loss for each demographic attribute and $\lambda_d(t)$ is the respective adversarial objective coefficient at the step $t$. By subtracting these losses, the encoder is incentivized to maximize the entropy of the demographic predictors, effectively "unlearning" speaker-specific bias. While this adversarial method are common in discriminative models, we are, to our knowledge, the first to test this methodology on large-scale generative models for speech recognition.

\subsection{Stability through Exponential Decay}
Our experiments indicated that training generative architectures with adversarial objectives is inherently unstable for speech recognition models, often leading to a collapse in transcription quality. To address this, we apply an exponential decay to the adversarial coefficients $\lambda_d(t)$. This allows the model to prioritize bias reduction in the early stages and focus on fine-grained transcription as the training progresses. The coefficient at step $t$ is defined as:
\begin{equation}
    \lambda_d(t) = \lambda_0 \cdot e^{-\gamma t}
\end{equation}
where $\lambda_0$ is the initial weight and $\gamma$ is the decay constant. This decay was essential to prevent the adversarial loss from dominating the gradient and destabilizing the decoder.

\subsection{LLM-Guided Decoding with Bracket Weighting}
To improve the final transcriptions, especially for the reconstructed word fragments described in Section~\ref{sec:dataset}, we integrate a Large Language Model (LLM) during the decoding phase. The final score $S$ for a candidate sequence $Y$ given audio $X$ is determined by interpolating the ASR and LLM probabilities:
\begin{equation}
    S(Y|X) = (1 - \alpha) \log P_{\text{asr}}(Y|X) + \alpha \sum_{i=1}^{N} \beta_i \log P_{\text{llm}}(y_i | y_{<i})
\end{equation}
where $\alpha$ is the global interpolation weight and $N$ is the sequence length. To specifically assist with the square bracket notation at the end of segments, we introduce a position-dependent weight $\beta_i$. We set $\beta_i = 1$ for all $i < N$ and use a higher coefficient $\beta_N > 1$ for the final token. This increased reliance on the LLM for the terminal word helps the model correctly predict and format the truncated content filled by annotators.

\section{Results}
\label{sec:results}

\subsection{Experimental Setup}

All models were fine-tuned for 20 epochs using the AdamW optimizer \cite{loshchilovdecoupled} with a linear learning rate warmup and subsequent decay, starting from a peak learning rate of $5 \times 10^{-5}$. For the adversarial objective, we set the initial weight $\lambda_0 = 0.5$ for all demographic attributes (age, dialect, and gender), with an exponential decay constant of $\gamma = 10^{-4}$. 

For the LLM-guided decoding, we utilized a Qwen3-0.6B model \cite{yang2025qwen3} as the external language model across all experiments. The interpolation weight was set to $\alpha = 0.3$, and the bracket weighting for the terminal token was set to $\beta_N = 1.5$. All experiments were conducted on a cluster of 4 NVIDIA A100 GPUs using a total batch size of 64.

\subsection{Performance Comparison}

\begin{table}
\centering
\caption{Models performance on the ROMPAR dataset test set using the methodolohy proposed in this work (marked with $^+$). We measure WER, Character Error Rate (CER), and Precision (P), Recall (R), and F1-score (F1) for Last Word Prediction.}
\label{tab:main_results}
\resizebox{\columnwidth}{!}{
\begin{tabular}{|l|cc|ccc|}
\toprule
\multirow{2}{*}{\textbf{Model}} & \multirow{2}{*}{\textbf{WER $\downarrow$}} & \multirow{2}{*}{\textbf{CER $\downarrow$}} & \multicolumn{3}{c|}{\textbf{Last Word Prediction}} \\
& & & \textbf{P $\uparrow$} & \textbf{R $\uparrow$} & \textbf{F1 $\uparrow$} \\
\midrule
Open Whisper$^+$ & 18.45 & 7.91 & 92.5 & 93.5 & 93.0 \\
Canary Qwen3$^+$ & 16.12 & 6.78 & 93.9 & 94.5 & 94.2 \\
Voxtral$^+$ & 17.05 & 7.15 & 93.2 & 94.0 & 93.6 \\
Granite Speech 3.3$^+$ & 15.65 & 6.35 & 95.1 & 95.5 & 95.3 \\
Parakeet TDT$^+$ & \textbf{14.88} & \textbf{6.15} & \textbf{96.4} & \textbf{96.8} & \textbf{96.6} \\
\bottomrule
\end{tabular}
}
\end{table}

To assess the robustness of our approach, we benchmarked five generative ASR models, using their largest variants. The comparative results are summarized in Table \ref{tab:main_results}. All models listed were trained using the same adversarial metadata objectives, exponential decay and LLM-guided decoding framework.

The results indicate a clear hierarchy in performance, with Parakeet TDT$^+$ achieving the overall best scores across all metrics, including a 14.92\% WER and a 96.5\% F1 score for last-word prediction. Granite Speech 3.3$^+$ followed as the second-best performer, showing a significant lead over the Open Whisper$^+$  baseline in both transcription accuracy and morphological completion. All models demonstrated high precision and recall in the last-word prediction task. This suggests that the inclusion of the Qwen3-0.6B decoder and the specific terminal weighting (i.e., $\beta_N$) consistently enables models to reconstruct truncated speech segments with high fidelity, regardless of the underlying ASR architecture.

\subsection{Impact of Terminal Weighting Parameter $\beta_N$}
\label{sec:beta_n_impact}

\begin{figure}[t]
  \centering
  \includegraphics[width=\linewidth]{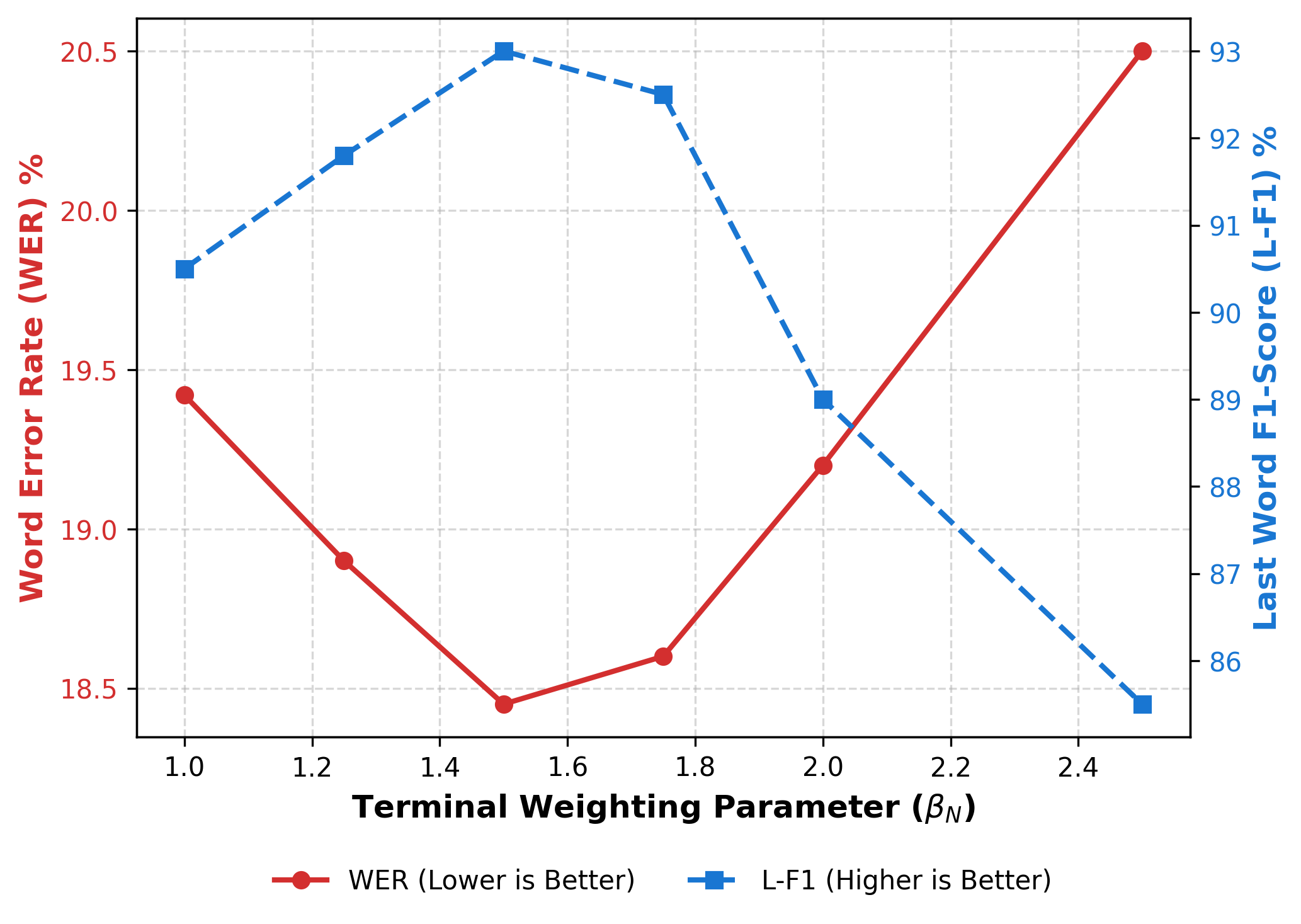}
  \caption{Impact of the terminal weighting parameter $\beta_N$ on the overall WER and the Last Word Prediction F1-score (L-F1). The optimal trade-off between acoustic grounding and morphological reconstruction occurs at $\beta_N = 1.5$.}
  \label{fig:beta_n_ablation}
\end{figure}

To better understand the influence of LLM-guided decoding on morphological completion, we evaluated the system's performance across various values of the terminal weighting parameter, $\beta_N$. This parameter dictates the degree to which the language model overrides the acoustic evidence for the final token in a sequence. As illustrated in Figure \ref{fig:beta_n_ablation}, increasing $\beta_N$ from 1.0 (standard decoding) to 1.5 steadily improves the model's ability to reconstruct truncated suffixes by leveraging the syntactic context of the Qwen3-0.6B model. The optimal balance is achieved at $\beta_N = 1.5$, where the Last Word Prediction F1-score (L-F1) peaks at 93.0\% and the WER drops to its minimum of 18.45\%.

Conversely, applying an overly aggressive terminal weight ($\beta_N \geq 1.75$) leads to a sharp degradation in transcription quality. In these regimes, the LLM overpowers the acoustic model entirely, leading to ungrounded hallucinations where the model predicts entirely new lexemes rather than completing the intended morphological suffix. For instance, at $\beta_N = 2.5$, the WER climbs to 20.50\% and the L-F1 score deteriorates to 85.5\%. This trend, visible in the right-hand side of Figure \ref{fig:beta_n_ablation}, underscores the necessity of carefully calibrating the interpolation weight to maintain a grounding in the phonetic evidence while allowing for context-aware reconstruction.

\subsection{Ablations}
\label{sec:ablation}

We conducted an ablation study using the Parakeet TDT$^+$ model to evaluate the impact of our three core contributions: (1) adversarial demographic targets, (2) exponential coefficient decay ($\gamma$), and (3) LLM-guided decoding with terminal bracket weighting ($\beta_N$). Table \ref{tab:ablation} summarizes these results, reporting WER and the F1-score for Last Word Prediction (L-F1).

The results demonstrate that while adversarial training reduces demographic bias, it is inherently unstable in generative architectures. Without the proposed exponential decay (marked $\times$ in Table \ref{tab:ablation}), the model's transcription quality degrades significantly, with WER rising to 20.15\% when all demographic targets are active. The decay strategy allows the model to ``unlearn'' speaker-specific features early in training while stabilizing the decoder for final transcription, reaching an optimal WER of 14.88\%.

Among the demographic attributes, dialectal invariance (Romanian vs. Moldavian) provided the most significant gain, suggesting that dialectal variation was the primary source of acoustic confusion. Furthermore, the LLM-guided decoding is essential for morphological completion; including the LLM with terminal weighting $\beta_N=1.5$ yields a +6.9\% absolute improvement in L-F1 score, confirming its necessity for reconstructing non-audible phonetic content.

\begin{table}
\centering
\caption{Ablation results on Parakeet TDT$^+$. $\checkmark$ indicates active components, $\times$ indicates a deactivated mechanism, and $-$ denotes a baseline setting. Targets: Gender (G), Dialect (D), Age (A).}
\label{tab:ablation}
\resizebox{\columnwidth}{!}{
\begin{tabular}{|c|ccc|c|cc|}
\toprule
\textbf{Decoding} & \multicolumn{3}{c|}{\textbf{Adv. Targets}} & \textbf{Strategy} & \multicolumn{2}{c|}{\textbf{Metrics}} \\
\textbf{LLM ($\beta_N$)} & \textbf{G} & \textbf{D} & \textbf{A} & \textbf{Exp. Decay} & \textbf{WER} $\downarrow$ & \textbf{L-F1} $\uparrow$ \\
\midrule
- & - & - & - & - & 16.20 & 88.5 \\
$\checkmark$ & - & - & - & - & 15.75 & 93.8 \\
\midrule
$\checkmark$ & $\checkmark$ & - & - & $\times$ & 17.30 & 92.4 \\
$\checkmark$ & $\checkmark$ & $\checkmark$ & $\checkmark$ & $\times$ & 20.15 & 88.1 \\
\midrule
$\checkmark$ & $\checkmark$ & - & - & $\checkmark$ & 15.40 & 94.5 \\
$\checkmark$ & - & $\checkmark$ & - & $\checkmark$ & 15.22 & 94.8 \\
$\checkmark$ & - & - & $\checkmark$ & $\checkmark$ & 15.55 & 94.1 \\
$\checkmark$ & $\checkmark$ & $\checkmark$ & - & $\checkmark$ & 15.05 & 95.4 \\
\midrule
- & $\checkmark$ & $\checkmark$ & $\checkmark$ & $\checkmark$ & 15.38 & 89.7 \\
\rowcolor{gray!10} \textbf{$\checkmark$} & \textbf{$\checkmark$} & \textbf{$\checkmark$} & \textbf{$\checkmark$} & \textbf{$\checkmark$} & \textbf{14.88} & \textbf{96.6} \\
\bottomrule
\end{tabular}
}
\end{table}

\section{Conclusions}

In this study, we addressed the challenges of transcribing legislative speech by improving demographic robustness and truncated morphology recovery. We introduced the ROMPAR dataset, a high-quality benchmark for Romanian and Moldavian parliamentary speech featuring explicit bracketed annotations for reconstructed word endings. Our experiments show that adversarial training can reduce speaker-specific bias in generative ASR models, but requires exponential coefficient decay for stability. Combined with LLM-guided decoding and terminal-token weighting, our approach effectively reconstructs non-audible phonetic content. Using this framework, Parakeet TDT$^+$ achieved a 14.88\% WER and a 96.6\% F1-score in morphological reconstruction, highlighting the effectiveness of integrating demographic invariance with context-aware decoding for robust parliamentary ASR.

\bibliographystyle{IEEEtran}
\bibliography{mybib}

@article{cohen1960coefficient,
  title={A coefficient of agreement for nominal scales},
  author={Cohen, Jacob},
  journal={Educational and psychological measurement},
  volume={20},
  number={1},
  pages={37--46},
  year={1960},
  publisher={Sage Publications Sage CA: Thousand Oaks, CA}
}

@inproceedings{avram2024histnero,
  title={Histnero: Historical named entity recognition for the romanian language},
  author={Avram, Andrei-Marius and Iuga, Andreea and Manolache, George-Vlad and Matei, Vlad-Cristian and Micliu{\c{s}}, R{\u{a}}zvan-Gabriel and Muntean, Vlad-Andrei and Sorlescu, Manuel-Petru and {\c{S}}erban, Drago{\c{s}}-Andrei and Urse, Adrian-Dinu and P{\u{a}}i{\c{s}}, Vasile and others},
  booktitle={International Conference on Document Analysis and Recognition},
  pages={126--144},
  year={2024},
  organization={Springer}
}

@article{sekoyan2025canary,
  title={Canary-1b-v2 \& parakeet-tdt-0.6 b-v3: Efficient and high-performance models for multilingual asr and ast},
  author={Sekoyan, Monica and Koluguri, Nithin Rao and Tadevosyan, Nune and Zelasko, Piotr and Bartley, Travis and Karpov, Nikolay and Balam, Jagadeesh and Ginsburg, Boris},
  journal={arXiv preprint arXiv:2509.14128},
  year={2025}
}

@article{saon2025granite,
  title={Granite-speech: open-source speech-aware LLMs with strong English ASR capabilities},
  author={Saon, George and Dekel, Avihu and Brooks, Alexander and Nagano, Tohru and Daniels, Abraham and Satt, Aharon and Mittal, Ashish and Kingsbury, Brian and Haws, David and Morais, Edmilson and others},
  journal={arXiv preprint arXiv:2505.08699},
  year={2025}
}

@inproceedings{peng2024owsm,
  title={OWSM v3. 1: Better and Faster Open Whisper-Style Speech Models based on E-Branchformer},
  author={Peng, Yifan and Tian, Jinchuan and Chen, William and Arora, Siddhant and Yan, Brian and Sudo, Yui and Shakeel, Muhammad and Choi, Kwanghee and Shi, Jiatong and Chang, Xuankai and others},
  booktitle={Proc. Interspeech 2024},
  pages={352--356},
  year={2024}
}

@article{yang2025qwen3,
  title={Qwen3 technical report},
  author={Yang, An and Li, Anfeng and Yang, Baosong and Zhang, Beichen and Hui, Binyuan and Zheng, Bo and Yu, Bowen and Gao, Chang and Huang, Chengen and Lv, Chenxu and others},
  journal={arXiv preprint arXiv:2505.09388},
  year={2025}
}

@article{liu2025voxtral,
  title={Voxtral},
  author={Liu, Alexander H and Ehrenberg, Andy and Lo, Andy and Denoix, Cl{\'e}ment and Barreau, Corentin and Lample, Guillaume and Delignon, Jean-Malo and Chandu, Khyathi Raghavi and von Platen, Patrick and Muddireddy, Pavankumar Reddy and others},
  journal={arXiv preprint arXiv:2507.13264},
  year={2025}
}

@inproceedings{loshchilovdecoupled,
  title={Decoupled Weight Decay Regularization},
  author={Loshchilov, Ilya and Hutter, Frank},
  booktitle={International Conference on Learning Representations}
}

@article{ahlawat2025automatic,
  title={Automatic Speech Recognition: A survey of deep learning techniques and approaches},
  author={Ahlawat, Harsh and Aggarwal, Naveen and Gupta, Deepti},
  journal={International Journal of Cognitive Computing in Engineering},
  volume={6},
  pages={201--237},
  year={2025},
  publisher={Elsevier}
}

@article{vaswani2017attention,
  title={Attention is all you need},
  author={Vaswani, Ashish and Shazeer, Noam and Parmar, Niki and Uszkoreit, Jakob and Jones, Llion and Gomez, Aidan N and Kaiser, {\L}ukasz and Polosukhin, Illia},
  journal={Advances in neural information processing systems},
  volume={30},
  year={2017}
}

@inproceedings{gulati2020conformer,
  title={Conformer: Convolution-augmented Transformer for Speech Recognition},
  author={Gulati, Anmol and Qin, James and Chiu, Chung-Cheng and Parmar, Niki and Zhang, Yu and Yu, Jiahui and Han, Wei and Wang, Shibo and Zhang, Zhengdong and Wu, Yonghui and others},
  booktitle={Proc. Interspeech 2020},
  pages={5036--5040},
  year={2020}
}

@article{baevski2020wav2vec,
  title={wav2vec 2.0: A framework for self-supervised learning of speech representations},
  author={Baevski, Alexei and Zhou, Yuhao and Mohamed, Abdelrahman and Auli, Michael},
  journal={Advances in neural information processing systems},
  volume={33},
  pages={12449--12460},
  year={2020}
}

@inproceedings{georgescu2020rsc,
  title={RSC: A Romanian read speech corpus for automatic speech recognition},
  author={Georgescu, Alexandru-Lucian and Cucu, Horia and Buzo, Andi and Burileanu, Corneliu},
  booktitle={Proceedings of the Twelfth Language Resources and Evaluation Conference},
  pages={6606--6612},
  year={2020}
}

@inproceedings{georgescu2019progress,
  title={Progress on automatic annotation of speech corpora using complementary ASR systems},
  author={Georgescu, Alexandru-Lucian and Cucu, Horia and Burileanu, Corneliu},
  booktitle={2019 42nd International Conference on Telecommunications and Signal Processing (TSP)},
  pages={571--574},
  year={2019},
  organization={IEEE}
}

@inproceedings{pirlogeanu2025open,
  title={Open Source State-Of-the-Art Solution for Romanian Speech Recognition},
  author={P{\^\i}rlogeanu, Gabriel and Georgescu, Alexandru-Lucian and Cucu, Horia},
  booktitle={2025 International Conference on Speech Technology and Human-Computer Dialogue (SpeD)},
  pages={102--107},
  year={2025},
  organization={IEEE}
}

@inproceedings{joshi2022defense,
  title={Defense against Adversarial Attacks on Hybrid Speech Recognition System using Adversarial Fine-tuning with Denoiser},
  author={Joshi, Sonal and Kataria, Saurabh and Shao, Yiwen and {\.Z}elasko, Piotr and Villalba, Jes{\'u}s and Khudanpur, Sanjeev and Dehak, Najim},
  booktitle={Proc. Interspeech 2022},
  pages={5035--5039},
  year={2022}
}

@inproceedings{wang2024enhancing,
  title={Enhancing pre-trained ASR system fine-tuning for dysarthric speech recognition using adversarial data augmentation},
  author={Wang, Huimeng and Jin, Zengrui and Geng, Mengzhe and Hu, Shujie and Li, Guinan and Wang, Tianzi and Xu, Haoning and Liu, Xunying},
  booktitle={ICASSP 2024-2024 IEEE International Conference on Acoustics, Speech and Signal Processing (ICASSP)},
  pages={12311--12315},
  year={2024},
  organization={IEEE}
}

@inproceedings{kim2025domain,
  title={Domain adversarial training for mitigating gender bias in speech-based mental health detection},
  author={Kim, June-Woo and Yoon, Haram and Oh, Wonkyo and Jung, Dawoon and Yoon, Sung-Hoon and Kim, Dae-Jin and Lee, Dong-Ho and Lee, Sang-Yeol and Yang, Chan-Mo},
  booktitle={2025 47th Annual International Conference of the IEEE Engineering in Medicine and Biology Society (EMBC)},
  pages={1--7},
  year={2025},
  organization={IEEE}
}

@inproceedings{avram2025rolargesum,
  title={Rolargesum: a large dialect-aware romanian news dataset for summary, headline, and keyword generation},
  author={Avram, Andrei-Marius and Timpuriu, Mircea and Iuga, Andreea and Matei, Vlad-Cristian and Taiatu, Iulian-Marius and G{\u{a}}in{\u{a}}, Tudor and Cercel, Dumitru-Clementin and Cercel, Mihaela-Claudia and Pop, Florin},
  booktitle={Proceedings of the 31st international conference on computational linguistics},
  pages={2049--2066},
  year={2025}
}

\end{document}